\documentclass[sigconf]{acmart}
\usepackage[ruled]{algorithm2e}
\renewcommand\footnotetextcopyrightpermission[1]{} 
\date{}
\usepackage{emptypage}

\AtBeginDocument{%
	\providecommand\BibTeX{{%
			\normalfont B\kern-0.5em{\scshape i\kern-0.25em b}\kern-0.8em\TeX}}}

\copyrightyear{2022}
\acmYear{2022}
\setcopyright{rightsretained}
\acmConference[]{Shuchao Deng}{Yanan Sun}{and, Edgar Galvan}
\acmDOI{10.1145/3520304.3528884}
\acmISBN{978-1-4503-9268-6/22/07}

\begin{document}

\title{Neural Architecture Search Using Genetic Algorithm for \\Facial Expression Recognition}

\author{Shuchao Deng}
\affiliation{%
	\institution{College of Computer Science, \\Sichuan University}
	\city{Chengdu}
	\country{China}}
\email{dengshuchao@stu.scu.edu.cn}

\author{Yanan Sun}
\authornote{Corresponding author.}
\affiliation{%
	\institution{College of Computer Science, \\Sichuan University}
	\city{Chengdu}
	\country{China}}
\email{ysun@scu.edu.cn}

\author{Edgar Galvan}
\affiliation{%
	\institution{Hamilton Institute, Dept. of CS, Maynooth University, Naturally Inspired Computation Res. Group}
	\country{Ireland}
}
\email{edgar.galvan@mu.ie}


\begin{abstract}
	Facial expression is one of the most powerful, natural, and universal signals for human beings to express emotional states and intentions. Thus, it is evident the importance of correct and innovative facial expression recognition (FER) approaches in Artificial Intelligence. The current common practice for FER is to correctly design convolutional neural networks' architectures (CNNs) using human expertise. However, finding a well-performing architecture is often a very tedious and error-prone process for deep learning researchers. Neural architecture search (NAS) is an area of growing interest as demonstrated by the large number of scientific works published in recent years thanks to the impressive results achieved in recent years.  We propose a genetic algorithm approach that uses an ingenious encoding-decoding mechanism that allows to automatically evolve CNNs on FER tasks attaining high accuracy classification rates. The experimental results demonstrate that the proposed algorithm achieves the best-known results on the CK+ and FERG datasets as well as competitive results on the JAFFE dataset.
\end{abstract}

%

\maketitle

\section{Introduction}

Facial expression recognition (FER)  has a variety of applications in human society, such as medical care, automotive, and robotics manufacturing \cite{bailly2017dynamic47}, to mention some. Convolutional neural networks (CNNs) are the most well-known networks thanks to their wide applicability in Euclidean data problems. These CNNs have become the standard architectures for FER tasks in multiple scientific works such as Deep-Emotion \cite{minaee2021deep05}, thanks to outperforming other non CNNs techniques. 

Architecture search is an area of growing interest as demonstrated by the large number of scientific works published in recent years and inspiring works have emerged for FER. For example, MnasNet-FER \cite{aghera2020mnasnet58} builds a recurrent neural network controller that continuously adjusts the input architecture. Auto-FERNet \cite{li2021auto36} weakens the search space to a continuous spatial structure and then combines the greedy algorithm. The ConvGP \cite{evans2018evolutionary60} uses genetic programming through a series of crossovers and mutations. These neural architecture search algorithms using CNNs on FER tasks perform better than those hand-crafted CNNs. However, there are some limitations on these works such as requiring high computational power as well as making strong assumptions on the search space by using fixed length representations. 

The main contribution of this work is addressing these issues. Specifically, 

\begin{itemize}
	\item We propose a variable-length encoding strategy to effectively address the need for a fixed-length encoding strategy.
	\item The skip connections are merged into the proposed algorithm to handle complex data. 
	\item A global caching system is set up to reduce the computational cost of the evolution process.
\end{itemize}

\section{Related Work}

\subsection{Neural Architecture Search}
Machine learning and deep learning are now being used in an increasing number of fields such as computer vision, healthcare, and robotics, thus requiring more and faster-automated design models. Google's proposal of NAS \cite{DBLP:journals/corr/ZophL16} caused a boom in the research community. Since then, NAS has attracted an increasing number of researchers due to its ability to automatically search for a good performing network \cite{sun2019evolving}. There are three main parts in NAS: search space, search strategy, and performance evaluation.

The search space defines which architectures can be represented. A search strategy details how to explore and exploit the search space, which is often exponentially large or even unbounded. The goal of performance estimation is usually to find architectures that achieve high classification performance for unseen data.

\subsection{Facial Expression Recognition}
Many researchers have started to explore the combination of NAS and CNN.  For example, Aghera et al. \cite{aghera2020mnasnet58}, proposed MnasNet-FER, an automatic mobile neural architecture-based approach for FER tasks. However, as pointed out by the authors, this approach is costly and difficult to balance between obtaining a lightweight architecture and obtaining a well-performing network. Li et al. \cite{li2021auto36}, proposed Auto-FERNet. This performs an automatic search of neural architectures through gradient optimization with pre-set layers of network architecture and construction of hypernets, which often requires a great deal of expertise. Another noteworthy method is the one proposed by Evans et al. \cite{evans2018evolutionary60}, dubbed ConvGP. This method overcomes the disadvantage of needing to pre-set the architecture length, but the excessive operators lead to requiring large computational calculations.

For FER tasks, we use a genetic algorithm approach to automatically search for network architecture, which is due to its extraordinary results in numerous areas and we employ ingenious ideas to overcome the issues of NAS on FER. 

\section{Methods}

\subsection{Algorithm Framework and Search Space}

\begin{figure}
	\centerline{\includegraphics[width=0.4\textwidth]{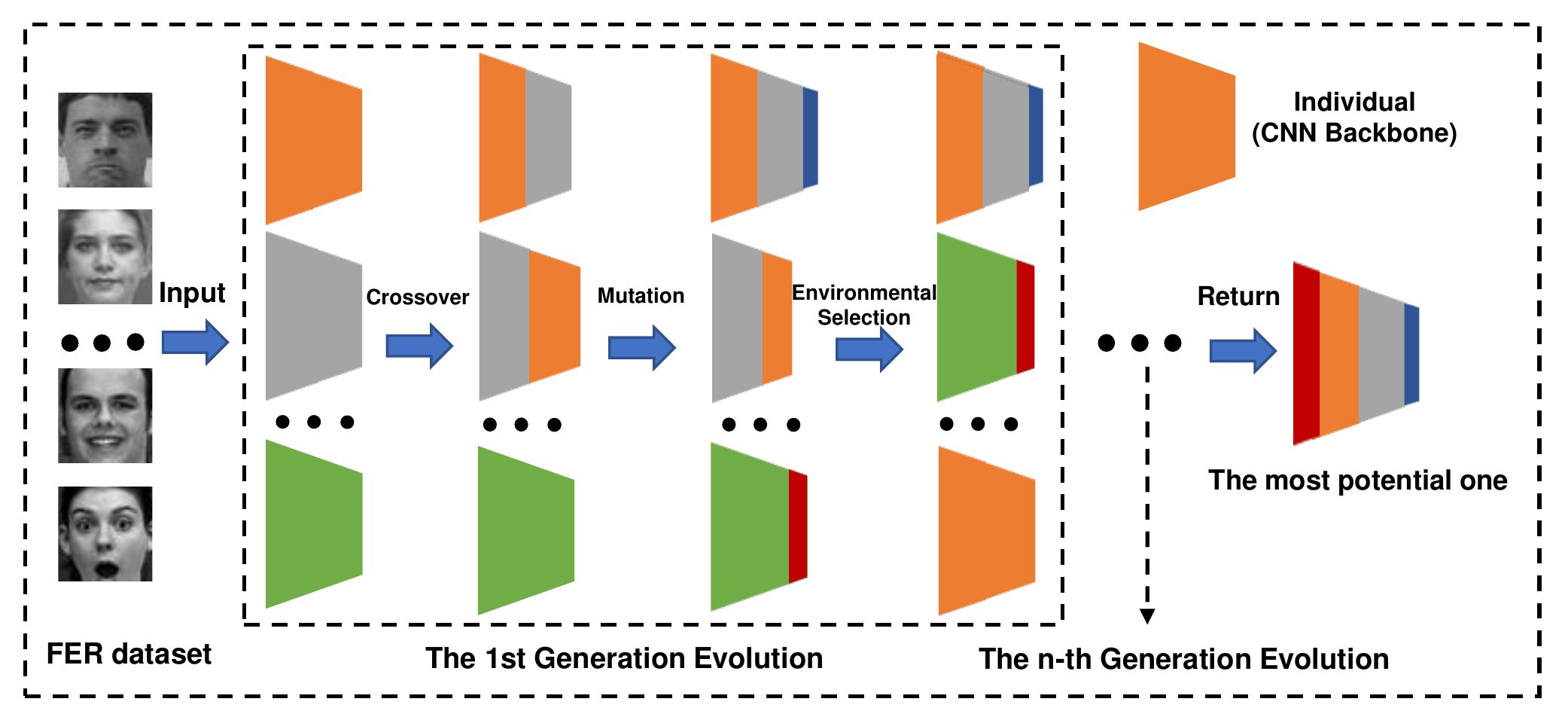}}
	\caption{The general flow of the proposed algorithm.}
	\label{fig3}
\end{figure}

Figure 1 demonstrates the overall framework of the proposed algorithm. The FER dataset is used as input, and through a series of evolutionary generations, the most potential CNNs’ architecture is automatically discovered. During the evolution process, firstly a population is randomly initialized to predefined population size, and the specific CNN architectures are encoded using the proposed encoding strategy. The fitness value of the individual is calculated, using accuracy as a metric. Then, the parent individuals are selected to generate new children according to the proposed crossover and mutation operators. Finally, the next generation of individuals is generated by selecting the parent and the new individuals, and the most potential one is returned.

The search space is defined by considering the following basic units: 3x3 \emph{convolution}, 2x2 \emph{maximum pooling}, 2x2 \emph{mean pooling} and \emph{skip connection}. The basic units form the basic blocks, where the skip connection layer consists of two convolutional layers and one skip connection. The proposed coding strategy is used to build the CNN architecture through these skip connections and pooling layers. The available numbers of feature maps are set to
{64, 128, 256, 512} based on the settings employed by the state-of-the-art CNNs, and the step size is set to 1x1 (inspired by the ResNet series). The pooling layer is divided into mean pooling and maximum pooling, and the step size is also set to 1x1. Figure 2 shows an example of a CNN architecture implemented using our variable-length coding strategy. 

\begin{figure*}
	\includegraphics[width=\textwidth]{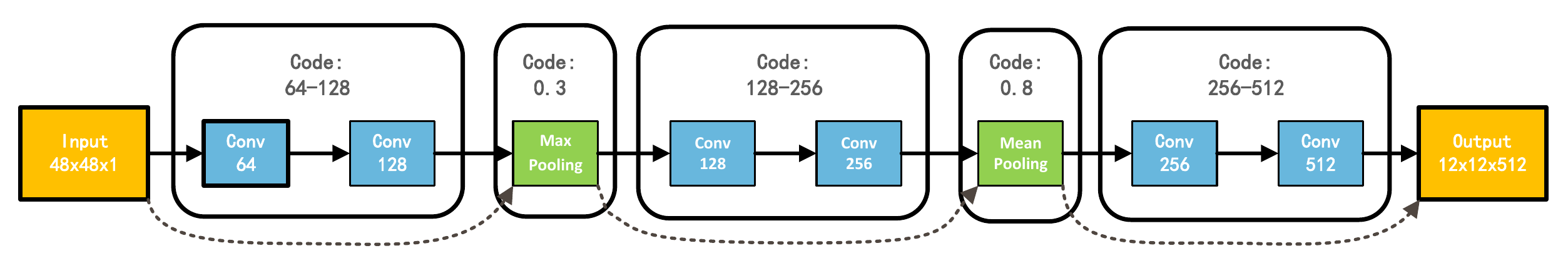}
	\caption{A network architecture formed with the proposed coding strategy.}
	\label{fig:teaser}
\end{figure*}

\subsection{Search Strategy and Performance Evaluation}

The first step is to initialize the population. Each individual in the population encodes the network architecture. Next, we use the acceleration component to compute fitness values for the population shown in Algorithm 1. We set up a global cache system for each individual (Lines 1-4), and if we find that the model corresponding to the individual is already in the global cache system, then we can get its fitness value without training and validation (Lines 6-9), and vice versa, we record it (Lines 10-19). Finally, we return the population with their corresponding fitness values (Line 22).

\begin{algorithm}[]
	\caption{ Fitness evaluation with global cache}
	\LinesNumbered 
	\KwIn{ The population $ P_t $ of the individuals, the FER dataset, the training epochs, the training data $ D_{train} $, the fitness evaluation data $ D_{val} $;}
	\KwOut{The population $ P_t $ with individuals' fitness values;}
	\If{$ t == 0 $}{
		$ Cache \leftarrow \phi $ \;
		Set $ Cache $ to a global variable\;
	}
	\ForEach{individual in $ P_t $ }{
		\eIf{the identifier of individual in Cache}{
			$ v \leftarrow $ Query the fitness by identifier from $ Cache $\;
			Set $ v $ to $ individual $\;
		}{
			$ v_{best} \leftarrow $ 0\;
			\ForEach{epoch in the given training epochs}{
				Train the CNN on $ D_{train} $\;
				$ v \leftarrow $\ Calculate the accuracy on $ D_{val} $\;
				\If{$ v > v_{best} $}{
					$ v_{best} \leftarrow v $ \;
				}
			}
			Set $ v_{best} $ as the fitness of $ individual $\;
			Put the identifier of $ individual $ and $ v_{best} $ into $ Cache $\;
		}
	}
	\Return $ P_t $;
\end{algorithm}

\begin{algorithm}
	\caption{Offspring generation}
	\LinesNumbered 
	\KwIn{ The population $ P_t $ containing individuals with fitness, the probability for crossover operation $ p_c $, the probability for mutation operation $ p_m $, the mutation operation list $ l_m $ , the probabilities of selecting different mutation operations $ p_l $ ;}
	\KwOut{The offspring population $ Q_t $;}
	$ Q_t \leftarrow \phi $ \;
	\While{$\left| Q_{t} \right| < \left| P_{t} \right|$}{
		$ p_1 \leftarrow $ Randomly select two individuals from $ P_{t} $, and from the two then select the better one\;
		$ p_2 \leftarrow $ Repeat Line 3\;
		\While{ $ p_2 == p_1 $ }{
			Repeat Line 4\;
		}
		$ r \leftarrow $ Randomly generate a number from (0,1)\;
		\eIf{$ r < p_c $}{
			Divide $ p_1 $ into two random parts by a point\;
			Divide $ p_2 $ into two random parts by a point\;
			$ o_1 \leftarrow $ The first part of $ p_1 $  adds the second part of $ p_2 $\;
			$ o_2 \leftarrow $ The first part of $ p_2 $  adds the second part of $ p_1 $\;
			$ Q_t \leftarrow Q_t \cup o_1 \cup o_2 $\;
		}{
			$ Q_t \leftarrow Q_t \cup p_1 \cup p_2 $\;
		}
	}
	\ForEach{individual $ p $ }{
		$ r \leftarrow $ Randomly generate a number from (0,1) \;
		\If{$ r < p_m $}{
			$ i \leftarrow $ Randomly choose a point in $ p $\;
			$ m \leftarrow $ Select one operation from $ l_m $ in $ p_l $\;
			Do the mutation $ m $ at the point $ i $ of $ p $\;
		}
	}
	Return $ Q_t $;
\end{algorithm}

 Algorithm 2 shows how crossover and mutation operators work. The first step in the algorithm is to select two parents using tournament selection of Size 2 (Lines 3-7). We designed single-point crossovers for variable-length coding for generating two offspring. (Lines 9-17). When mutating an individual, a specific mutation operation is selected from the provided mutation list (Lines 19-26). In the proposed algorithm, the available mutation operations defined in the mutation list are:
\begin{itemize}
	\item adding a skip connection layer with random settings; 
	\item adding a pooling layer with random settings; 
	\item remove the layer at the selected location; and 
	\item randomly changing the parameter values in the selected location building block.
\end{itemize}
The first two mutation operators increase the network depth and the third mutation operator decreases the network depth. Crossover and mutation followed by return of offspring (Line 27).

Finally, there is an environmental selection to form the next generation of individuals, and we use a binary tournament selection, as mentioned before, as well as elitism. 

\section{Experiments}
\subsection{Benchmark Datasets}

The Extended Cohn-Kanade (CK+) dataset \cite{lucey2010extended} contains 593 video sequences from a total of 123 different subjects, ranging from 18 to 50 years of age with a variety of genders and heritage. We divided the dataset into training set, validation set and test set, the numbers of which are 687, 101 and 193 respectively. The JAFFE dataset \cite{lyons2020coding,lyons2021excavating} consists of 213 images of different facial expressions from 10 different Japanese female subjects. We used 120 images for training, 23 images for validation, and 70 images for the test (10 images per emotion in the test set). FERG \cite{aneja2017modeling} is a database of cartoon characters with annotated facial expressions containing 55,769 annotated face images of six characters. We use around 34k images for training, 14k for validation, and 7k for testing.

\subsection{Parameter Settings}
We set the population size to 20 and the evolutionary generations to 20. The crossover and mutation probabilities in the parameter settings for this algorithm to search for the most potential CNN architecture are set to 0.9 and 0.2. We trained a total of 600 epochs using stochastic gradient descent with an initial learning rate of 0.025, a momentum of 0.9, and a weight decay of 3e-4 and adjusted the learning rate to 0.017 at 100 epochs, 0.001 at 300 epochs, and 0.0001 at 500 epochs. In addition, the number of available feature maps is set to 64,128, 256, 512 according to the settings used in state-of-the-art CNN. For the FERG dataset, we set the epochs to 20 and keep the rest of the values the same. The algorithm can be stopped when the evolutionary generation is over or the performance is good. The results of all experiments were averaged over five tests.

\subsection{Experimental Results}

To show the effectiveness and efficiency of the proposed algorithm, we selected the state-of-the-art algorithms as peer competitors.

\begin{table}[]
	\centering
	\footnotesize
	\caption{Comparison of accuracy with peer competitors.}\label{tab:1}
	\setlength{\tabcolsep}{0.5mm}
	\begin{tabular}{cccccc}
		\hline
		\textbf{Method} & \textbf{Pre-train} & \textbf{CK+} & \textbf{JAFFE} & \textbf{FERG} & \textbf{Manual or Auto} \\ \hline
		Deep-Emotion \cite{minaee2021deep05} & No & 98.00 & 92.80 & 99.30 & Manual  \\ 
		Ensemble Multi-feature \cite{zhao2018transfer65} & No & - & 80 & 97 & Manual  \\ 
		Adversarial NN \cite{feutry2018learning} & Yes & ~ & - & 98.2 & Manual  \\ 
		LBP \cite{kola2021novel67} & No & 93.9 & 88.3 & 96.7 & Manual  \\
		SAFER \cite{yaddaden2018user46} & No & 96.37 & 95.30 & - & Manual  \\ 
		FAN \cite{meng2019frame40} & Yes & 99.69 & - & - & Manual  \\ 
		AFER \cite{yaddaden2021facial44} & No & - & 96.05 & - & Manual  \\ 
		FERIK \cite{cui2020knowledge39} & Yes & 97.59 & - & - & Manual  \\ 
		HMTL \cite{DBLP:journals/corr/abs-2105-06421} & Yes & 98.23 & 79.88 & - & Manual  \\
		ViT \cite{Aouayeb.202177} & Yes & 98.17 & 94.83 & - & Manual  \\ 
		ViT + SE \cite{Aouayeb.202177} & Yes & 99.80 & 92.92 & - & Manual  \\ 
		ViT \cite{Aouayeb.202177} & No & 98.57 & 88.23 & - & Manual  \\ 
		ViT + SE \cite{Aouayeb.202177} & No & 99.49 & 90.61 & - & Manual  \\ 
		Auto-FERNet \cite{li2021auto36} & No & 98.89 & \textbf{97.14} & - & Auto  \\ 
		ConvGP \cite{evans2018evolutionary60} & No & - & 96.67 & - & Auto  \\ \hline
		Ours & No & \textbf{100} & 95.71 & \textbf{99.98} & Auto  \\ \hline
	\end{tabular}
\end{table}

 As shown in Table 1, we achieved 100\% accuracy on the CK+ dataset and 99.98\% accuracy on the FERG dataset, both achieving the best-known results so far, and 95.71\% accuracy on the JAFFE dataset, higher than most models, achieving competitive results. Furthermore, compared to manual design, our algorithm is completely free of manual design. In particular, in contrast to the currently existing methods that perform well only on a single dataset, our proposed algorithm performs well on multiple datasets, thus demonstrating the good adaptability of our proposed algorithm.

\subsection{Analysis of Results}
There are three key aspects that these modern and revolutionary methods have focused their attention on: size of the network, the number of parameters and the time to train these networks. These three elements are shown in Table 2.

As it can be seen, the number of layers varies for each of the datasets used in this work. This shows how the proposed mutation operators work effectively to automatically adjust the size of the network, with 16, 12 and 13 layers for the CK+, JAFFE and FERG datasets, respectively. The number of parameters is also reported in this table, third column from left to right. Finally, we show how it is possible to effectively train our network without requiring massive computational power. In this work, we use an NVIDIA 2080 Ti GPU card. We can see that the time, measured in GPU-hours, goes from 16 to +27 hours, for the CK and the other two datasets, respectively. These times show how our proposed encoding is effective on FER tasks, where it has been well documented that multiple high-end GPUs are required to train these types of networks within reasonable time \cite{galvan2021neuroevolution}.

\begin{table}[]
	\caption{The number of layers and parameters of the network architecture and the time spent in the architecture search.}\label{tab:2}
	\begin{tabular}{cccc}
		\hline
		\textbf{Dataset} & \textbf{Layers} & \textbf{Parameters (M)} & \textbf{Time (GPUhs)} \\ \hline
		CK+              & 16              & 10.2                & 29                   \\
		JAFFE            & 12              & 5.9                 & 27.75                \\
		FERG             & 13              & 11.07               & 16                   \\ \hline
	\end{tabular}
\end{table}

\section{Conclusion and Future Work}

We propose an algorithm for automatically designing network architectures based on genetic algorithms for FER. Specifically, we propose a variable-length encoding strategy and the corresponding crossover operator to efficiently explore the optimal network depth. Secondly, skip connections are introduced into the algorithm to make it possible to handle complex data. Finally, a global caching system is set up to reduce the computational cost of the evolution process. Experimental results show that our algorithm achieves good performance. In the future, we will work on accelerated fitness evaluation methods to apply the proposed algorithms to larger datasets.

\bibliographystyle{ACM-Reference-Format}
\bibliography{reference}

\appendix

\end{document}